\documentclass[sigconf]{acmart}

\usepackage{booktabs} 

\renewcommand\footnotetextcopyrightpermission[1]{} 
\setcopyright{rightsretained}

\begin{document}
	\title{COTA: Improving the Speed and Accuracy of Customer Support through Ranking and Deep Networks}
	
	\author{Piero Molino}
	\affiliation{%
		\institution{Uber AI Labs}
		\city{San Francisco}
		\state{California}
		\postcode{94103}
	}
	\email{piero@uber.com}
	
	\author{Huaixiu Zheng}
	\affiliation{%
		\institution{Uber Technologies}
		\city{San Francisco}
		\state{California}
		\postcode{94103}
	}
	\email{huaixiu.zheng@uber.com}
	
	\author{Yi-Chia Wang}
	\affiliation{%
		\institution{Uber Technologies}
		\city{San Francisco}
		\state{California}
		\postcode{94103}
	}
	\email{yichia.wang@uber.com}
	
	\begin{abstract}
For a company looking to provide delightful user experiences, it is of paramount importance to take care of any customer issues.
This paper proposes COTA, a system to improve speed and reliability of customer support for end users through automated ticket classification and answers selection for support representatives.
Two machine learning and natural language processing techniques are demonstrated: one relying on feature engineering (COTA v1) and the other exploiting raw signals through deep learning architectures (COTA v2).
COTA v1 employs a new approach that converts the multi-classification task into a ranking problem, demonstrating significantly better performance in the case of thousands of classes.
For COTA v2, we propose an Encoder-Combiner-Decoder, a novel deep learning architecture that allows for heterogeneous input and output feature types and injection of prior knowledge through network architecture choices.
This paper compares these models and their variants on the task of ticket classification and answer selection, showing model COTA v2 outperforms COTA v1, and analyzes their inner workings and shortcomings.
Finally, an A/B test is conducted in a production setting validating the real-world impact of COTA in reducing issue resolution time by 10 percent without reducing customer satisfaction. 
\end{abstract}

	%

	
	\maketitle
	\section{Introduction}

Customer support has become an integral part of most companies.
In fact, many companies have customer service departments with dedicated representatives/agents to provide support and help customers resolve issues they encounter.
Prompt and accurate response to a request is essential for customer satisfaction and retention.
Uber\textquotesingle s Customer Obsession team is committed to making the customer experience as quick, easy, and accessible as possible.

When customers report problems, it is important to route them to the best possible resolution in a timely manner. 
Nevertheless, there is often a large amount of situational information associated with each customer ticket, which can be time-consuming to digest and synthesize into an identifiable issue type and corresponding solution for customer support representatives (CSRs).
Moreover, the diversity of ways a customer can describe an issue associated with a ticket further complicates the ticket resolution process.
Finally, as a company scales, support agents must be able to handle an ever-increasing volume and diversity of support tickets.
For example, Uber receives hundreds of thousands of tickets every day on the platform across 400+ cities worldwide.
The issues can range from technical errors and fare adjustments, to lost items. 
Ensuring that agents are empowered to resolve tickets as accurately and quickly as possible is the key.

Prior work has explored the challenge of issue resolution utilizing data mining and machine learning techniques to partially automate the resolution process \citep[e.g.,][]{DBLP:journals/ci/GuptaGF13,DBLP:journals/taslp/GuptaTHBRG06,DBLP:journals/iam/HuiJ00,DBLP:conf/cods/MuniRCVB17}.
Hui and Jha  \cite{DBLP:journals/iam/HuiJ00} applied data mining methods to extract customer-related information from both unstructured text data and structured databases, which CSRs employ for decision making.
In \cite{DBLP:journals/ci/GuptaGF13}, the authors built a ML classifier to identify emotional emails sent by customers and suggested that the classifier can be used to automatically route these emails to specialized representatives. 
Similarly, existing research in spoken dialogue systems aims to build models to detect intent and extract named entities for call classification and routing \citep[e.g.,][]{DBLP:journals/taslp/GuptaTHBRG06, DBLP:conf/asru/XuS13, IEEE:5947649}.  
 
In contrast to prior work, which mostly focuses on customer information retrieval and intent detection for routing, this work investigates techniques to directly help CSRs improve their speed and accuracy, which in turn leads to better customer experiences.
To accomplish this, we build COTA (Customer Obsession Ticket Assistant), an intelligent system based on machine learning (ML) and natural language processing (NLP) techniques that is integrated with Uber's customer support platform.
The system provides CSRs with suggested ticket classifications and answers based on ticket content and additional context such as relevant trip information.

In this paper, we share our experiences building COTA and report empirical results after successfully integrating it with Uber\textquotesingle s customer support platform.
The main contributions of this work are as follows:

\begin{itemize}
\item Proposing a new method to convert a multi-classification problem to a ranking one, which significantly improves the performance of feature-engineered classical ML algorithms, especially when there are thousands of classes.  The new method is introduced in Section~\ref{cota_v1}.
\item Introducing Encoder-Combiner-Decoder, a novel deep learning architecture that allows for heterogeneous input and output types and the injection of prior knowledge in the form of architecture choices.  The new architecture is introduced in Section~\ref{cota_v2}.
\item Conducting comprehensive experiments to compare different models and uncover their underlying mechanisms and shortcomings.  Experiments are discussed in Section~\ref{experiments_and_results}.
\item A/B testing COTA in production to show that it enables quick and efficient issue resolution and improves key business metrics.  Business impact is discussed in Section~\ref{business_impact}.
\end{itemize}

\section{Intelligent System for CSRs}

As one of the world's largest ridesharing providers, Uber receives hundreds of thousands of support tickets from users every day.
Until COTA, the process of resolving a ticket has been mostly manual.
A typical workflow for ticket resolution involves two steps: \textbf{contact type identification} and \textbf{solution selection}.  

When a CSR opens a ticket, their first step is to determine what it is about (e.g.\ \textit{wrong food delivered} or \textit{trip cancellation}).
We refer to this task as \textbf{contact type identification} (similar to  intent detection in dialogue systems research).
With thousands of potential contact types and a deep tree to navigate through, reducing the amount of time a CSR spends identifying a ticket\textquotesingle s type is important because it also decreases the time customers have to wait for their issue to be solved. 

Once a contact type is chosen, the next step is to select the right solution and reply.
Specifically, a customer service team usually maintains a bank of \textbf{reply templates} from which agents can select the correct one for each contact.
This step of finding the proper solution and selecting the appropriate reply is also time consuming since there are thousands of possible solutions to choose from and each contact type has a different set of protocols and solutions it is associated with.
We refer this task as \textbf{reply template selection}.

\subsection{COTA System Architecture}

\begin{figure}
\includegraphics[width=0.7\columnwidth]{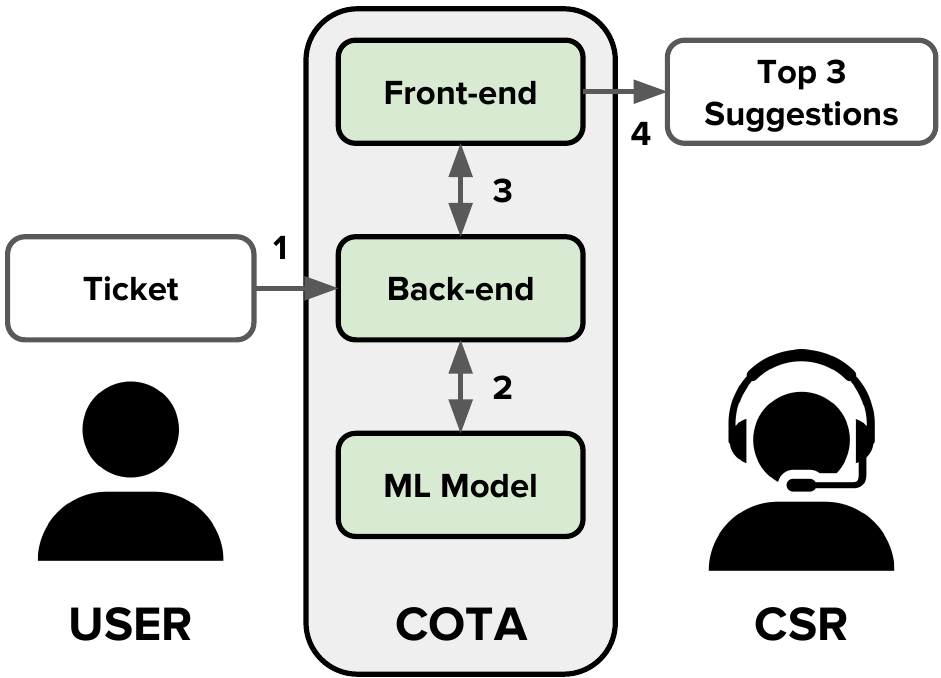}
\caption{The COTA system architecture is composed of a four-step workflow.}
\label{fig:cota_architecture}
\end{figure}

To solve both contact type identification and reply template selection, Customer Obsession Ticket Assistant (COTA) is designed to help our CSRs improve their speed and accuracy.
Built on top of our support platform, COTA is comprised of two ML models that suggest the three most likely contact types and reply templates to CSRs for each support tickets based on its content and context.
These two models are the \textbf{type model} and the \textbf{reply model}, respectively. 
Fig.~\ref{fig:cota_architecture} depicts the general COTA architecture, which follows a four-step flow:

\begin{enumerate}
\item Once a new ticket is issued by the user, the back-end service collects all relevant features of the contact.
\item The back-end service then sends these features to the ML models, receives back predictions and stores them.
\item Once a CSR opens a given ticket, the front-end service triggers the back-end service to check if there are any updates to the ticket. If there are no updates, the back-end service will return the stored predictions; if there are updates, it will fetch the updated features and go through Step 2 again.
\item The top three ranked predictions are suggested to agents; from there, agents make a selection and resolve the support ticket. The decision of returning the top three is a UX decision.
\end{enumerate}

\subsection{Model Features}
 
There are four main sources of information often referenced by CSRs when deciding contact type and reply template of a ticket: ticket message and metadata, user-level as well as trip-level information.
These sources of user and trip information contain critical ingredients for contact type identification and reply template selection, and as such serve as features for training the model. 

\begin{description}
\item[Ticket message.] The text written by the customer when submitting a ticket.  A detailed description of how ticket messages are processed and converted into features through an NLP pipeline is given in Section~\ref{nlp_preprocessing}.
\item[Ticket metadata.] Every support ticket comes with a set of metadata, such as \textit{ticket creation time} and \textit{product type} (e.g., Uber Eats, UberPool and UberX). 
\item[User information.] Information about the user sending the ticket, i.e. user type (e.g., driver, rider or eater), can provide important signals for both contact type identification and reply template selection.  For instance, only eaters would submit a ticket related to the contact type \textit{wrong food delivered}, while the country of the user may affect the policies to solve the tickets, and consequently the appropriate reply template to use.
\item[Trip information.] For tickets related to trips, trip-level information can be very helpful in predicting contact types, such as \textit{cancellation} and \textit{mistimed trips}.  Examples of trip features are \textit{city}, \textit{estimated time of arrival} and \textit{trip status}.  
\end{description}

\section{COTA v1: Traditional Machine Learning Models} \label{cota_v1}

\begin{figure}
\includegraphics[width=\columnwidth]{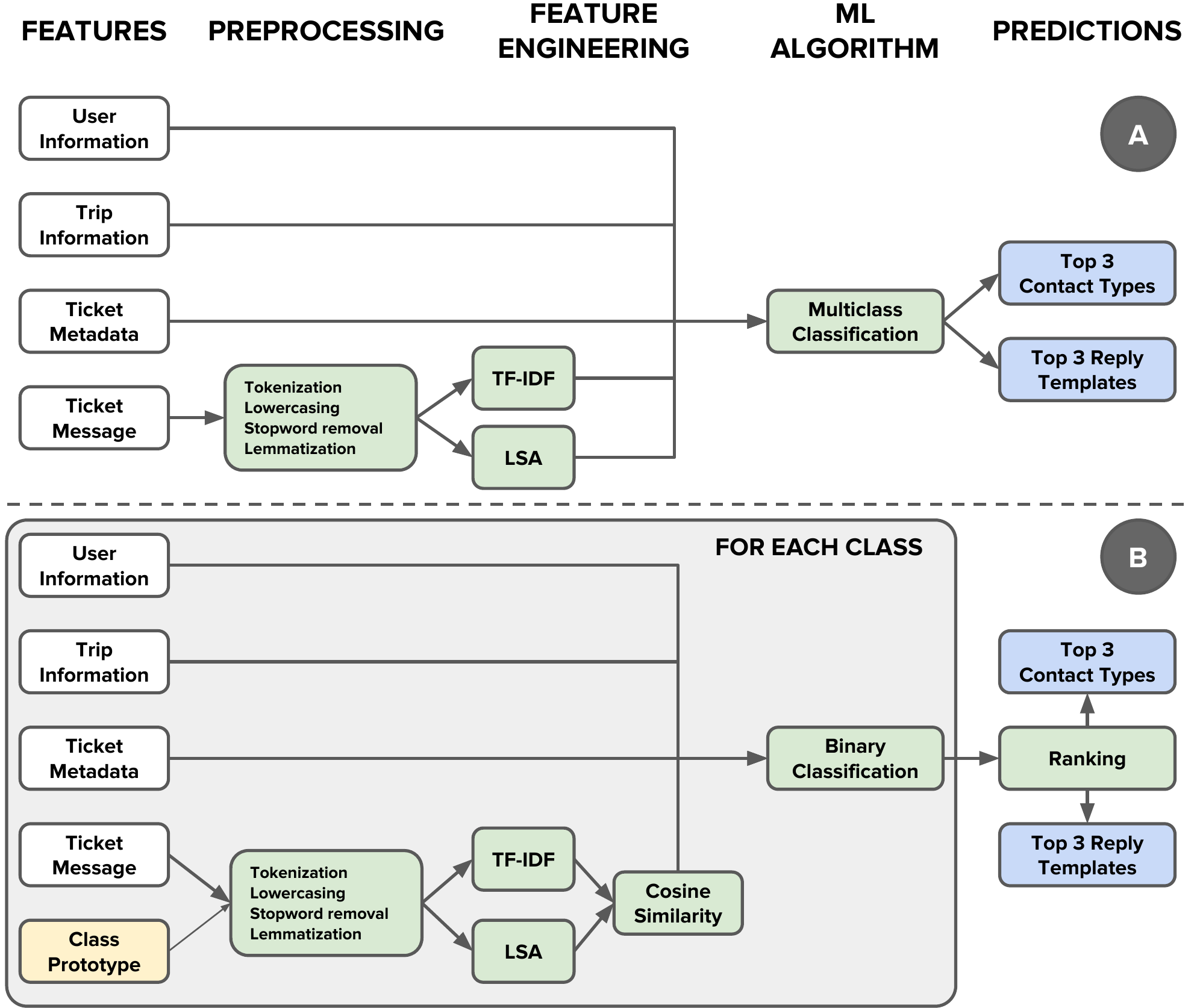}
\caption{The NLP pipeline built for COTA v1: a) topic vectors are directly used by the classification algorithm and b) cosine-similarity features are engineered and used by the pointwise-ranking algorithm.}
\label{fig:mcc_vs_rnk}
\end{figure}

The first version of COTA (COTA v1) is built with topic-modeling-based traditional NLP and ML techniques leveraging a mixture of text, categorical, and numerical features.
In order to extract the text features, an NLP pipeline processes incoming ticket messages.
This section describes each component of the pipeline shown in Fig.~\ref{fig:mcc_vs_rnk}.

\subsection{NLP Preprocessing} \label{nlp_preprocessing}
The first step is to analyze text at the word-level and use topic modeling to better understand the meaning of text data.
The text is cleaned by removing HTML tags.
Next, the message's sentences are tokenized and stop-words are removed.
Then, each word is lemmatized to convert different inflected forms into the same base form.
Finally, the documents are converted into a bag of words, forming a dictionary.

\begin{figure}
\includegraphics[width=0.47\columnwidth]{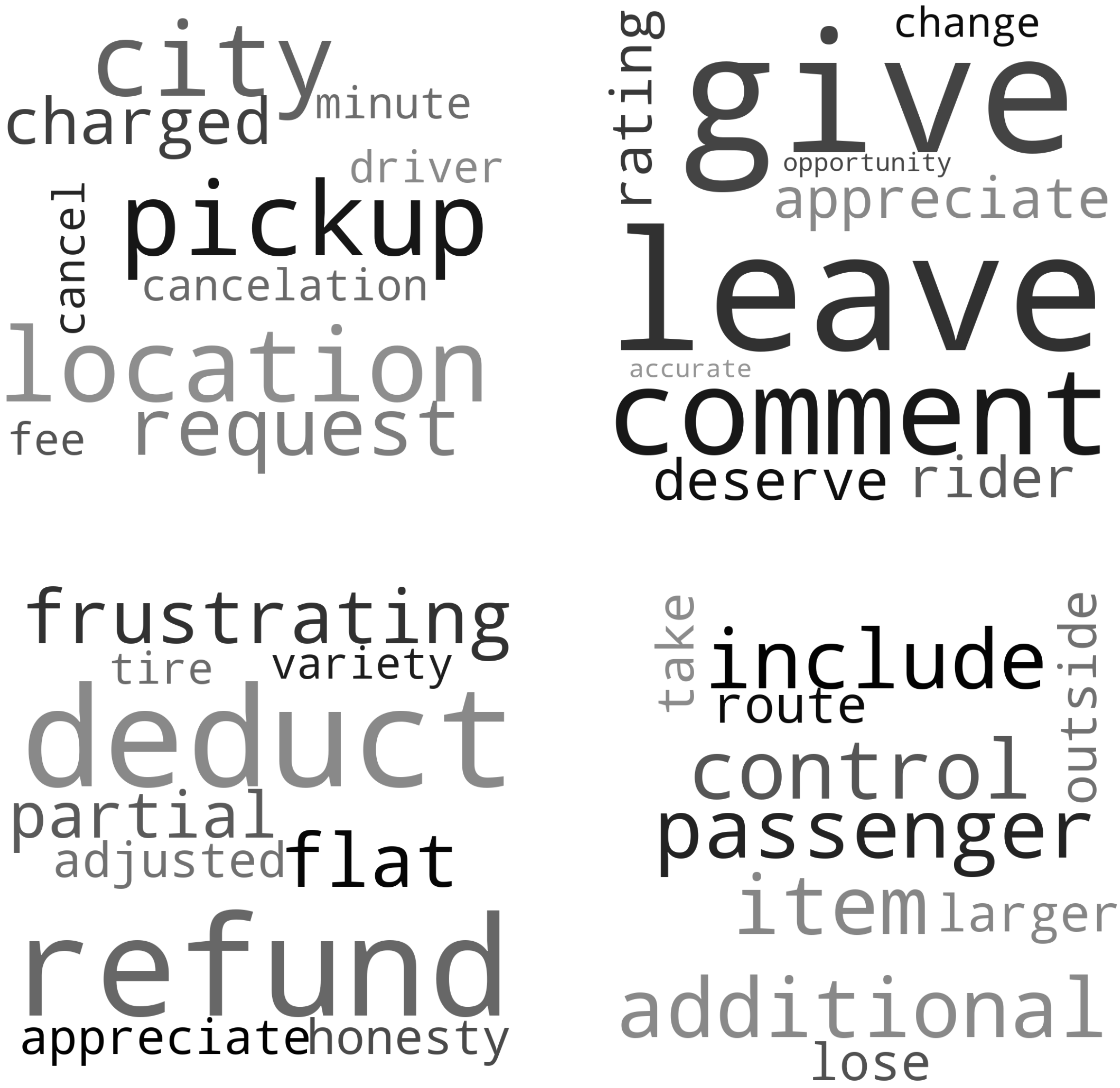}
\caption{Examples of topics extracted by LSA from text data in customer support tickets.}
\label{fig:topics}
\end{figure}

To understand user intent, topic modeling is performed on the bag of words after preprocessing.
First, TF-IDF (term frequency-inverse document frequency)~\cite{DBLP:conf/ike/McCullohDC08} is used to obtain a sparse vector representation.  Furthermore, topics are extracted through Latent Semantic Analysis (LSA)~\cite{DBLP:conf/cscl/LandauerD02}, that returns a dense vector for each ticket message.
We perform Truncated-SVD, that decomposes our term-document matrix $T$ obtained from ticket messages into $T \approx U_k\Sigma_k V_k^\top$, choosing the number of largest singular values $k$ in the $\Sigma$ matrix by keeping at least 90\% of the variance of the term-document matrix.
By doing so, we end up with about $200$ dimensions.
Fig.~\ref{fig:topics} shows examples of the topics extracted by LSA, and they are meaningful topics with respect to Uber's customer support ticket data, e.g.\ city related topics, rating related ones, refunds, fare adjustments and so on.

\subsection{Feature Engineering}
Two different approaches of using topic-modeling-based vector representations are tested.

The first approach is to directly leverage the topic vectors of ticket messages as features to perform downstream classifications in type and reply models, as shown in Fig.~\ref{fig:mcc_vs_rnk}(A). However, this direct approach suffers from a high dimensionality of the vectors. 

The second approach uses the vectors in an indirect fashion by performing further feature engineering by computing cosine similarity features, as illustrated in Fig.~\ref{fig:mcc_vs_rnk}(B).
Historical tickets associated with each contact type and reply template are collected and a bag-of-words representation is obtained for each of them. 
The bag-of-words representation of each class $i$ (either contact type and reply template) is transformed into a LSA vector, which we use as a prototype vector $p_i$.
The bag of words of each incoming ticket message $j$ is projected in the same semantic space in order to obtain a vector $t_j$.
The cosine similarity score $s_{ij}$ between each $p_i$ and $t_j$ is then computed and represents the similarity between class $i$ and ticket $j$.
The same process is repeated with TF-IDF vectors in place of LSA ones.
Doing so reduces the feature space from hundreds of dimensions of the original vectors to just a handful similarity features.

Similarity feature collection is further expanded by using other class specific information explicitly.
For example, in order to obtain an additional prototype, we can use the actual textual content of the reply template.
This adds two additional similarity features, one obtained with TF-IDF vectors and the other one with LSA ones.

\subsection{Algorithms: Multi-Class Classification vs. Pointwise-Ranking}

Given the two approaches to extract features from text described above, two different algorithms are trained on the tasks of contact type identification and reply template selection.

The first one formulates the tasks as a multi-class classification problem where the contact types and reply templates are targets. The model takes in TF-IDF and LSA vectors together with the categorical and numerical features of the other feature families and use them to predict the target class.
The pipeline for this straightforward approach is shown in Fig.~\ref{fig:mcc_vs_rnk}(A).

In order to leverage the engineered cosine-similarity features, a pointwise-ranking ~\cite{DBLP:conf/cikm/LeiLLZ17} algorithm (Fig.~\ref{fig:mcc_vs_rnk}(B)) scores each ticket-class pair and then ranks classes based on the score.
Specifically, a subset of all ticket-class pairs is sampled: the matching one is given a positive label (1) and a random subset of non-matching ones are given a negative label (0).
The target to predict is the label $Y_{ij}$ (0/1) for each pair of class $i$ and ticket $j$.
Using the cosine similarity features as well as categorical and numerical features, a binary classification algorithm is built to classify whether or not each ticket-class combination matches.
Once the algorithm scores each possible pair, classes are ranked based on their scores and the top-ranked ones are selected.

In COTA v1, both the multi-class classification and point-wise ranking algorithms use the same RandomForest algorithm~\cite{DBLP:journals/pami/Ho98} for learning to predict the correct class of each ticket.

The two approaches are compared in an experiment described in Section~\ref{exp_mcc_vs_rnk}.

\section{COTA v2: Deep Learning Architecture}  \label{cota_v2}

In recent years, deep learning models have been heavily adopted for several NLP tasks including syntactic~\cite{DBLP:conf/acl/BowmanGRGMP16} and semantic parsing~\cite{DBLP:conf/acl/LiangBLFL17}, semantic role labeling~\cite{DBLP:conf/emnlp/MarcheggianiT17}, recognizing textual entailment~\cite{conf/iclr/RocktaschelGHKB16} and named entity recognition~\cite{DBLP:journals/jmlr/CollobertWBKKK11}, leading to performance improvements.
Moreover, the same architectures employed for these tasks have been adopted also for end-to-end applications like summarization~\cite{DBLP:conf/naacl/ChopraAR16}, machine translation~\cite{conf/iclr/BahdanauCB15}, building dialogue systems~\cite{DBLP:conf/interspeech/Hakkani-TurTCCG16} and, more importantly for our goals, text categorization~\cite{DBLP:conf/emnlp/Kim14}.

Trying to leverage these models is a natural step for COTA, as the tasks to automate can be casted as text classification. In \cite{DBLP:conf/nips/ZhangZL15} the authors have already shown how deep learning architectures can be effective in the presence of big datasets.
The main difference with respect to state-of-the-art text classification architectures, is that they are explicitly designed to deal with textual inputs, while additional input features such as metadata, trip as well as user related information are available in COTA.
To incorporate these extensive features, COTA v2 extends the wide-and-deep approach described in \cite{/conf/dlrs/Cheng16}.
The wide-and-deep approach consists in combining other features with the text ones by concatenating them before the final layer of the architecture.
COTA v2 combines this idea with multi-task learning, that accommodates ways to train a model to learn to predict multiple outputs by optimizing losses for all the different outputs at the same time.
In \cite{DBLP:journals/jmlr/CollobertWBKKK11}, the authors showed how training models for related tasks in this fashion can lead to improved performances in all tasks.

\subsection{Encoder-Combiner-Decoder Architecture}

\begin{figure}
\includegraphics[width=\columnwidth]{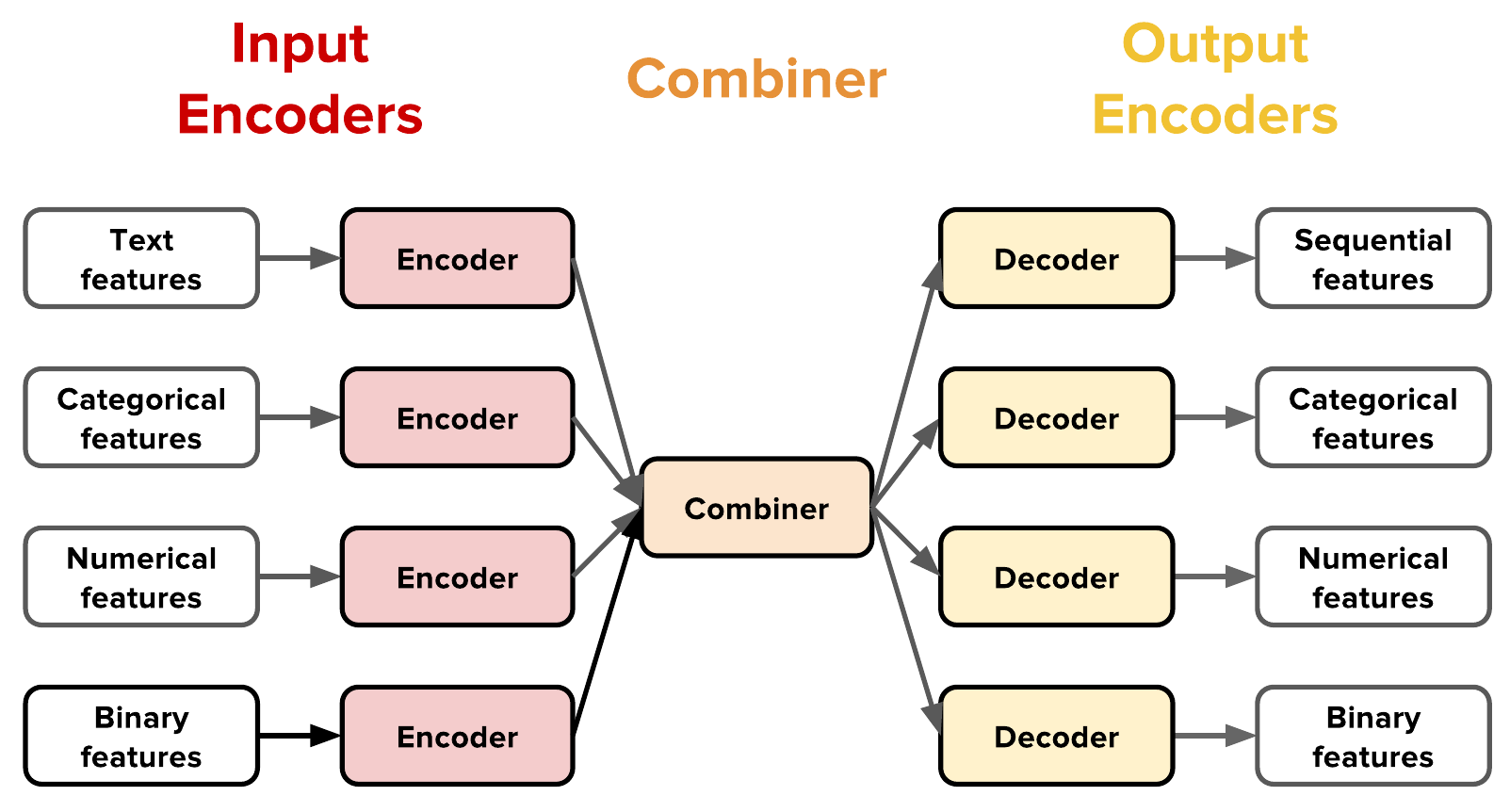}
\caption{Encoder-Combiner-Decoder Architecture depiction.}
\label{fig:cotav2_ecd}
\end{figure}

The combination of deep-and-wide with multi-task learning inspired losses forms the basis of a new general architecture introduced in this paper, the Encoder-Combiner-Decoder (ECD), depicted in Fig.~\ref{fig:cotav2_ecd}.
In the encoder part of the architecture, each single input feature is encoded by a sub-part of the model, depending on its type.
More formally each encoder is a function $e_{t(\phi_i(x))}(\phi_i(x))$ where $\phi_i(x)$ is the $i$-th input feature and $t(\phi(x))$ is the type of feature.
For instance, text features can by encoded by a Character-Based CNN or by a Bidirectional RNN on the word sequence, categorical features can be encoded through a linear projection into an embedding space, binary feature can be encoded with one single number and numerical features can be encoded through a single neuron that acts as a learned scaling factor.
Each of these different encoders outputs a vector encoding for the input feature they deal with.
In the combiner part, those vectors are concatenated as they are in the wide-and-deep approach, but the concatenation is optionally followed by fully connected layers that can learn some non-linear combination of the representations obtained so far.
The combiner is needed in any circumstance when there is more than one input feature, like in our case, but could be skipped if there's more there is only one input feature.
A combiner is defined as a function $c(x)$ so that:
$$c(x)=f([e_{t(\phi_0(x))}(\phi_0(x)), \dots, e_{t(\phi_n(x))}(\phi_n(x))])$$,
where $[\dots]$ is the concatenation operator, $n$ is the number of input features, and $f$ is a multi-layer perceptron.
Finally, in the decoder part, different output features have different ``heads'', with each of them predicting a different output feature depending on their type.
Each output decoder can contain an arbitrary number of additional fully connected layers between the output of the combiner and the layer responsible for the prediction. This makes it possible to have a multi-task model with weights shared among all the tasks up to the combiner and have a set of weights that are task-specific for increased flexibility.
Each decoder is a function $d$ that returns a loss and is defined as:
$$d_{\phi_i(x)}(x)=f_{\phi_i(x)}(c(x))$$,
where $f$ is again a multi layer perceptron and there could be different $f$s for different output features.
Categorical output features are treated as a multi-class classification task: they use the output of the combiner and pass it through softmax layer, and return a categorical cross entropy loss.
Numeric output features are treated as a regression task: they use the output of the combiner to predict a single value, and return a mean squared error loss. 
Binary output features are treated as a binary classification task: they use the output of the combiner as input for a layer with a logistic activation, and return a binary cross entropy loss.
The sum of all the losses coming from the different output features is optimized through any variant of stochastic gradient descent in multi-task learning fashion.
The final loss that is optimize is the weighted sum of all the losses of the output features and is defined as:
$$L(x)=\sum_{i \in o(x)}w_i d_{\phi_i(x)}(x)$$,
where $o(x)$ is the set of indices of output features, and $w_i$ is a user defined weight for the specific output feature.

The ECD architecture provides three key benefits. It directly incorporates all raw input features, eschewing the need for preprocessing outside of mapping textual and categorical features into integers, as well as learning a single model to predict both contact types and reply templates.
Finally, the architecture enables an easy method for swapping and comparing different encoders.
As a result, six diferent encoders are implemented for textual features: CNN, RNN and CNN followed by RNN, each one working on characters or on words, inspired by \cite{DBLP:conf/emnlp/Kim14, DBLP:conf/nips/ZhangZL15, DBLP:conf/acl/TaiSM15}.
Each of them can be parametrized independently, specifying the number of layers, the size of convolutional filters of each layer and how many stacked RNNs to use, if they should be bidirectional, and their type (Simple RNN~\cite{DBLP:journals/cogsci/Elman90}, LSTM~\cite{DBLP:journals/neco/HochreiterS97} or GRU~\cite{DBLP:conf/emnlp/ChoMGBBSB14}) among many other hyperparameters.

We are going to release \textbf{ludwig}, an open source toolbox implementing the ECD architecture.

\subsection{Injecting Prior Knowledge in the Architecture} \label{injecting_prior_knowledge_architecture}

The ECD architecture makes it possible to further improve model performance by injecting prior knowledge about each task directly at the model architecture level.

\subsubsection{Predicting paths in a tree}

\begin{figure}
\includegraphics[width=0.6\columnwidth]{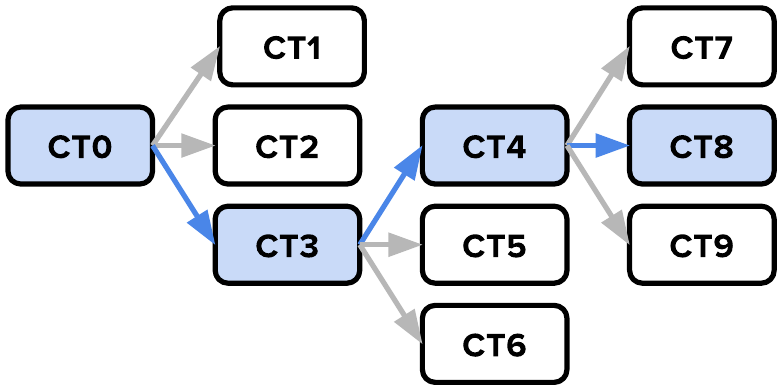}
\caption{Example of path in the contact type hierarchy used as target sequence. CT stands for contact type. The contact type to predict is CT8 and the target sequence is [CT0, CT3, CT4, CT8].}
\label{fig:cotav2_cth}
\end{figure}

\begin{figure}
\includegraphics[width=0.85\columnwidth]{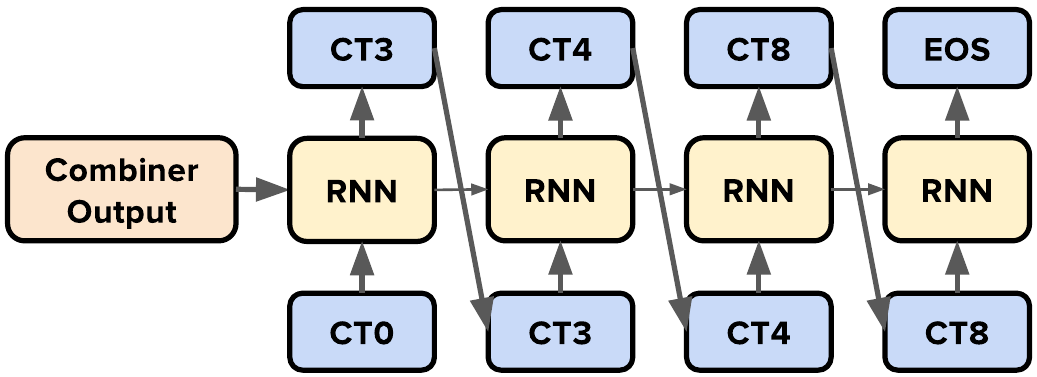}
\caption{Example of a sequence decoder predicting a path in the contact type tree.}
\label{fig:cotav2_sd}
\end{figure}

The first way to inject prior knowledge deals with the decoder for predicting contact types.
As previously described, contact types are organized in a hierarchy, but are treated as independent classes in COTA v1.
In COTA v2 the hierarchical nature of the contact types is exploited directly by predicting a path in the tree that leads to the target contact type rather than predicting the contact type directly, as shown in Fig.~\ref{fig:cotav2_cth}.
This is accomplished by adding an additional sequential decoder, similar to what is used in sequence-to-sequence models~\cite{DBLP:conf/nips/SutskeverVL14} for machine translation, where the target sequence is the sequence of contact type nodes that describes a path in the tree from the root to the target contact type, such that the last node in the sequence is the actual target contact type.
The node prediction at each step is used as input at the following step of sequence prediction, as depicted in Fig.~\ref{fig:cotav2_sd}.
The loss for this output feature is the sum of the cross entropy losses at each step of the sequence generation, while only the last predicted contact type in the sequence is used when assessing its accuracy.

Interestingly, not all model mistakes are equal in terms of time it takes for CSRs to recover from errors.
In fact, different model architectures may gracefully degrade and provide help for CSRs even if the answer is not exactly correct.
For instance predicting the parent of the correct contact type is better than predicting a contact type in a completely different part of the hierarchy because the CSRs will be just one click away from the correct solution.
In contrast, navigating through all the tree to select the correct contact type consumes a considerably larger amount of time.
Moreover, searching for solutions using a beam search could even improve handling time, providing an efficient approximation of the $k$ most likely paths rather than the most likely single path.
The hypothesis that a recurrent decoder could make more reasonable mistakes is directly tested in Section~\ref{testing_prior_knowledge_injection_hypotheses}.

\subsubsection{Adding decoder dependencies}

\begin{figure}
\includegraphics[width=\columnwidth]{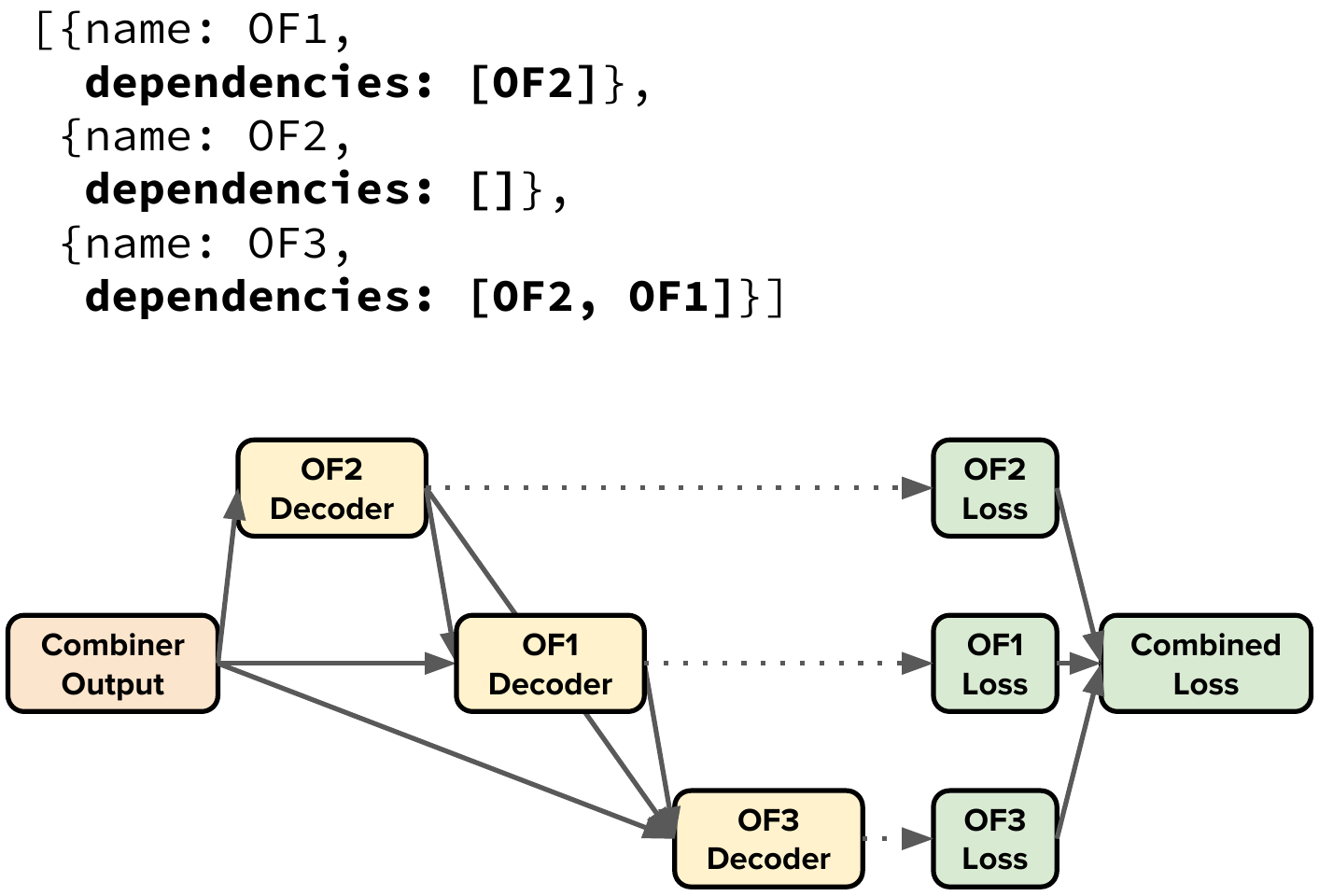}
\caption{Decoder dependency example. OF stands for output feature.}
\label{fig:cotav2_ofd}
\end{figure}

The second way to inject prior knowledge deals with the logical dependency between the two tasks we try to solve.
Contact type identification is logically the first step that CSRs undergo in the process of solving a ticket, and depending on the outcome, they decide what reply template to use. 
From the model perspective, having access to information about the contact type of the ticket is beneficial for suggesting the correct reply template CSRs should use.

As different decoders can have a set of weights that is specific to output feature they try to predict, the output of the contact type decoder is injected as an additional input to the reply template decoder, concatenating it with the output of the combiner.
More generally, in the ECD architecture allows for dependencies between output features.
The dependencies must form a directed acyclic graph for the computational graph to be built as when building it the outputs of the decoders of all the output features that the current output feature is dependent on is concatenated with the output of the combiner.
This concatenation is used as the input to the current decoder, as exemplified in Fig.~\ref{fig:cotav2_ofd}.

The hypothesis is that adding dependencies among output features would help in two ways: a) the overall performance in predicting output features with dependencies should improve due to to the prior knowledge injection and b) the outputs of the decoders will be more coherent with each other, i.e.\ the number of samples where all output predictions are correct as opposed to the number of samples where only a subset of output predictions is correct should be higher, even in the case of equivalent performance on each single output prediction.
Both hypotheses are tested in Section~\ref{testing_prior_knowledge_injection_hypotheses}.

\section{Experiments and Results} \label{experiments_and_results}

\subsection{Goals and Evaluation Metrics}

The goals of the experiments are twofold: to evaluate both the quality of the predictions obtained from these models and also how the use of the models impacts business metrics.

For the first goal, we evaluate models in an offline experiment in terms of their Accuracy and Hits$@$3 on both contact type identification and reply template selection.
The decision to use Hits$@$3 as a metric is rooted in the UX decision to show the top three model predictions to CSRs, so the number correct predictions we can provide among the ones actually shown to our users is assessed.
Moreover, the models should to return correct predictions on both tasks at the same time, so the accuracy of predicting the correct contact type and the correct reply template for the same tickets is considered. 

The business metrics that relate directly to how quickly and seamlessly we are able to resolve customer support issues, thereby improving the customer support experience, are the most important ones.
The goal is to provide them with a fast customer support experience that solves their issues.
For this reason, the handling time of each ticket and overall customer satisfaction is tracked through surveys.
The hypothesis is that when using COTA, CSRs will be faster in handling tickets without degrading customer satisfaction.
This hypothesis is tested with an online experiment, where A/B testing is performed on a subset of our English-speaking CSRs, using as treatment the intelligent suggestions generated by our COTA system in contrast to control with no suggestions.

\subsection{Dataset}

Uber's historical customer support data is used to train and evaluate the model performance.
As mentioned before, the input features include ticket message, ticket metadata, user-level information, and trip-level information.
The target variable for the task of contact type identification is the contact type ID, while for the task of reply template selection, the predicted contact type is used as an additional input, and the target variable is the reply template ID.
About $3$M tickets were collected and split randomly into train ($2.8$M), validation ($94$K) and test ($94$K) sets.
The dataset contains thousands of different contact types and reply templates, exhibiting a long-tail distribution.
Contact types in particular are structured in seven levels deep hierarchy.
Ticket messages are truncated to 1024 characters, as only 0.001\% of them being longer and the majority of them concentrating around 250 characters.

\subsection{Hyperparameters}

To find the best hyperparameters for each model, a hyperparameter search is conducted, selecting hyperparameter combinations based on model accuracy for a validation set.

For COTA v1 models, a grid search on the hyperparameters of the Random Forest was run to find the following optimal set : estimators: 100, max depth: 100, max features: $sqrt(\#features)$, min samples leaf: 50.

\begin{table}
  \caption{Performance on the validation set of the best hyperparameter setting for each model using different text encoders. The reported minutes are relative to a full validation set evaluation. \textit{CT} stands for contact type, \textit{RT} stands for reply template, \textit{Comb.} stands for Combined.}
  \label{tab:hyperopt}
  \begin{tabular}{rcccc}
    Encoder & CT Acc & RT Acc & Comb. Acc & Min\\
    \midrule
	Char C-RNN & \textbf{0.6505} & \textbf{0.5155} & \textbf{0.5024} & 17.5 \\
	Word CNN   & 0.6433 & 0.5099 & 0.4960 & \textbf{2} \\
	Word RNN   & 0.6413 & 0.5096 & 0.4971 & 8.5 \\
	Word C-RNN & 0.6315 & 0.4999 & 0.4841 & 6 \\
	Char CNN   & 0.6298 & 0.4990 & 0.4815 & 2.5 \\
	Char RNN   & 0.6262 & 0.4972 & 0.4767 & 36 \\
\end{tabular}
\end{table}

For COTA v2 models, the hyperparameter search is more complex, as different text encoders needed a completely different set of hyperparameters.
A parallel random search is performed, cutting at 100 different configurations for each different encoder architecture.
Performance and speed are reported in Table~\ref{tab:hyperopt}.
There is a relatively small performance spread between the best and worst hyperparameter combinations of each encoder, with the worst performing encoder (Char RNN) obtaining just 2\% lower accuracy than the best performing (Char C-RNN).
In the end, Word CNN was chosen as text encoder as it performs less than 1\% worse than the Char C-RNN, while being approximatively 9 times faster during both training and prediction.

The best Word CNN model has 256 dimensional word embeddings and 4 parallel 1D convolutional layers of size, respectively, 2, 3, 4 and 5 with 512 filters each.
All categorical features were embedded with 256 dimensional embeddings.
The combiner does not have fully connected layers, while the two decoders had two fully connected layers of size 512 and 256 respectively.
Each numeric feature is encoded using only a batch norm~\cite{DBLP:conf/icml/IoffeS15} layer. 
Fully connected layers interleaved by dropout layers with a dropout probability of 0.35.
The loss is optimized with Adam~\cite{conf/iclr/KingmaB14} using a learning rate of 0.00025 and a batch size of 256.

\subsection{COTA v1: Classification vs. Ranking} \label{exp_mcc_vs_rnk}

Experiments on COTA v1 compare multi-class classification and pointwise-ranking algorithms on the same dataset and the same optimized hyperparameters.
The results are shown in Table~\ref{tab:cl_vs_rk}. 

The ranking algorithm outperforms the classification algorithm significantly on both type identification and reply selection, as measured by the metrics of accuracy, Hits@3 and combined accuracy of the two tasks. 
In particular, for the contact type identification, the ranking improves accuracy by more than ${\sim}3\%$ and Hits$@$3 by ${\sim}6\%$.
For the reply template selection, the improvement by obtained by ranking algorithm is much more pronounced: ${\sim}11\%$ on accuracy and ${\sim}19\%$ on Hits$@$3.
This can be attributed to the fact that in the ranking framework for reply template selection, the meta text information (the content) of reply templates is injected directly into the cosine-similarity feature engineering.
As expected, it significantly boosts the model performance compared to classification. As a results of the improvements to both type and reply models, ranking algorithm has a much stronger performance on the overall accuracy of both tasks: ${\sim}14\%$ more accurate compared to classification.
These empirical results highlight the importance of feature engineering in cases where a non-deep-learning algorithm is employed to tackle the task at hand.

\begin{table}
  \caption{Comparison between Classification and Ranking in COTA v1.}
  \label{tab:cl_vs_rk}
  \begin{tabular}{rccccc}
     & \multicolumn{2}{c}{Contact Type} & \multicolumn{2}{c}{Reply Template} & Combined \\
     & Acc & Hits$@$3 & Acc & Hits$@$3 & Accuracy \\
    \midrule
    Classification & 0.4583 & 0.6118 & 0.3669 & 0.5315 & 0.3190 \\
    Ranking & \textbf{0.4913} & \textbf{0.6716} & \textbf{0.4753} & \textbf{0.7207} & \textbf{0.4552} \\
\end{tabular}
\end{table}

\subsection{COTA v2: Testing Prior Knowledge Injection Hypotheses} \label{testing_prior_knowledge_injection_hypotheses}

The three hypotheses regarding the knowledge injection in the model architecture discussed in Section~\ref{injecting_prior_knowledge_architecture} are tested:

\begin{enumerate}
\item The sequential decoder could learn to make more reasonable mistakes.
\item Adding a dependency from contact type to reply template should improve performance in predicting reply templates.
\item Adding a dependency from contact type to reply template should improve performance in predicting both contact type and reply templates at the same time.
\end{enumerate}

\begin{table}
  \caption{Categorical decoder (multi-class contact type classification) vs Sequential decoder (predicts paths in the contact type tree) on contact type prediction. \textit{Accuracy+p} indicates accuracy considering also the parent of the target node as a correct prediction.}
  \label{tab:cat_cs_seq}
  \begin{tabular}{rcc}
    Contact Type Decoder & Accuracy & Accuracy+p\\
    \midrule

    Categorical & 0.6321 & 0.6743 \\ 
    Sequential & \textbf{0.6568} & \textbf{0.7342} \\
\end{tabular}
\end{table}

Hypothesis 1 is confirmed by the results shown in Table~\ref{tab:cat_cs_seq}.
The sequential decoder is slightly more accurate than the categorical one.
Furthermore, parents of the correct contact types are included as correct predictions (\textit{Accuracy+p}), as those are considered reasonable mistakes, and the sequential decoder obtains a ${\sim}6\%$ higher score in this setting, confirming that it makes more reasonable mistakes than the categorical one.

\begin{table}
  \caption{Effect of adding dependency between the reply template decoder and the contact type decoder.}
  \label{tab:add_dep}
  \begin{tabular}{rccc}
     & \multicolumn{2}{c}{Reply Template} & Combined \\
     Reply Template Decoder & Accuracy & Hits$@$3 & Accuracy \\
    \midrule
    No dependency & 0.5061 & 0.7042 & 0.4012 \\ 
    With dependency & \textbf{0.5471} & \textbf{0.7521} & \textbf{0.5312} \\
\end{tabular}
\end{table}

Hypothesis 2 and 3 are confirmed by the results shown in Table~\ref{tab:add_dep}.
Adding the dependency between the reply template decoder and the contact type one improves the performance on the reply template prediction by ${\sim}4\%$ accuracy and approximate ${\sim}5\%$ Hits$@$3.
The biggest improvement is visible on the combined accuracy of both prediction tasks, where adding the dependency improves the accuracy by ${\sim}13\%$ and brings the score close to the accuracy on the reply template alone, confirming that the dependency helps coherence between both outputs.

\subsection{Comparison between COTA v1 and v2}

\begin{table}
  \caption{Comparison between COTA v1 and COTA v2.}
  \label{tab:v1_vs_v2}
  \begin{tabular}{rccccc}
     & \multicolumn{2}{c}{Contact Type} & \multicolumn{2}{c}{Reply Template} & Combined \\
     & Acc & Hits$@$3 & Acc & Hits$@$3 & Accuracy \\
    \midrule
    COTA v1 & 0.4913 & 0.6716 & 0.4753 & 0.7207 & 0.4552 \\ 
    COTA v2 & \textbf{0.6568} & \textbf{0.7258} & \textbf{0.5471} & \textbf{0.7521} & \textbf{0.5312} \\
\end{tabular}
\end{table}

After obtaining the best models for COTA v1 and COTA v2, both models are compared directly for all tasks measuring the combined accuracy.
The results in Table~\ref{tab:v1_vs_v2} show how COTA v2 outperforms COTA v1 in all tasks with a variable margin, most notably in contact type accuracy by ${\sim}16\%$ and in combined accuracy by ${\sim}8\%$ .
This result is in line with literature on text classification where deep learning models outperform other ML approaches when big amounts of training data is available.

\subsection{Model Analysis of COTA v2}

\begin{figure*}[t!]
\includegraphics[width=0.92\linewidth]{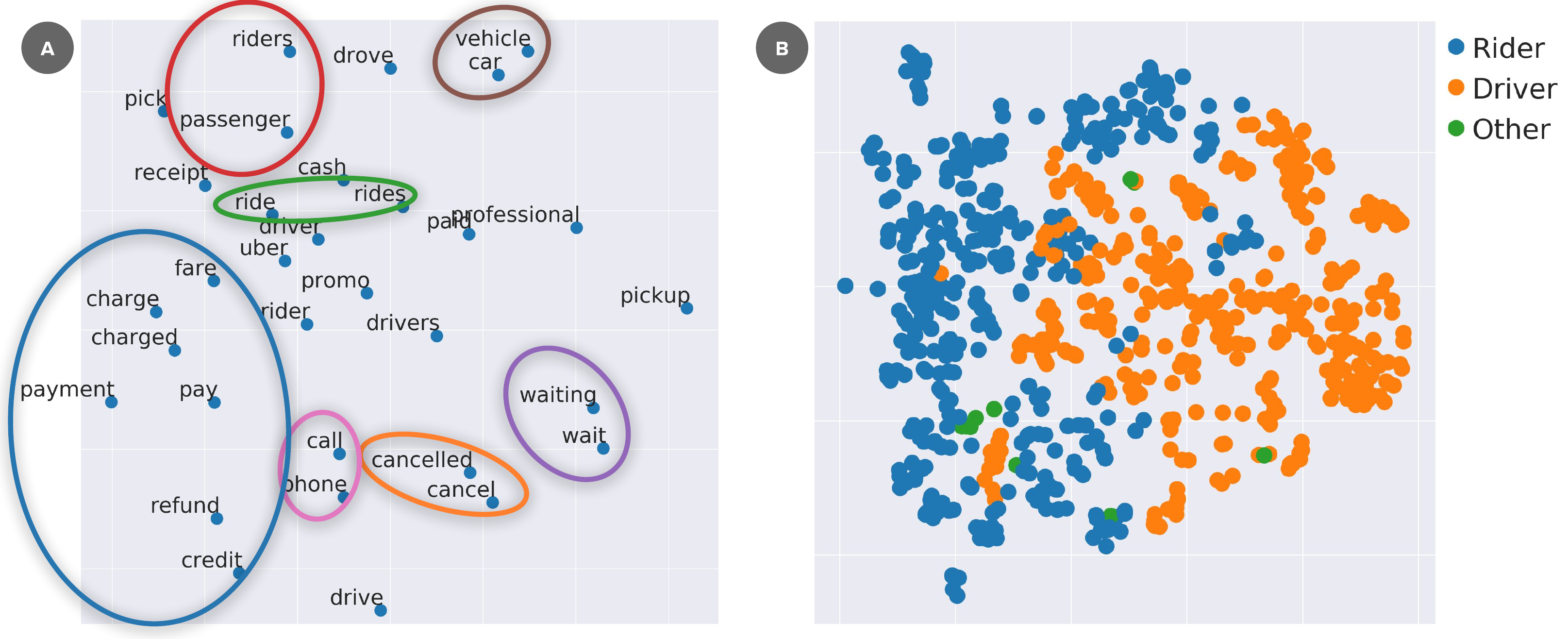}
\caption{Embeddings learned by the deep learning models: a) word embeddings, b) entity embeddings of contact types.}
\label{fig:cotav2_embeddings}
\end{figure*}

Here, the deep learning models in COTA v2 are analyzed in order to understand how it works and its error modes.
The representations learned as a byproduct of training on the impact of class imbalance on model performance are visualized.

To gain insight into the model inner working, the embeddings the model learns for the words in the tickets and for the contact types are visualized.
The high-dimensional embeddings space is projected to 2d using t-SNE~\cite{vanDerMaaten2008}.
Fig.~\ref{fig:cotav2_embeddings}(A) shows the word embeddings of a set of keywords often encountered in COTA's use case. Meaningful clusters emerge in the the t-SNE plot.
Semantically related words such as ``car'' and ``vehicle'', ``phone'' and ``call'' appear to be close to each other.
Fig.~\ref{fig:cotav2_embeddings}(B) shows the embeddings learned for the contact types with each data point corresponding to one unique contact type. The embeddings are extracted from the weights of the last fully-connected layer before the softmax layer in the deep learning model for classifying contact types. Each column of the weight matrix can be interpreted as the embedding encoding of a contact type class.
The contact types are color coded into three major groups, namely ``rider'', ``driver'', and ``other'' (e.g., eater, restaurant, etc.).
The t-SNE plot shows clear clustering of rider and driver related contact types.
These visualizations intuitively confirm that the model is learning reasonable representations and suggest that the model is capable to capture correlations and semantical connections between words and the relationship between contact types. 

\begin{figure}
\includegraphics[width=0.85\columnwidth]{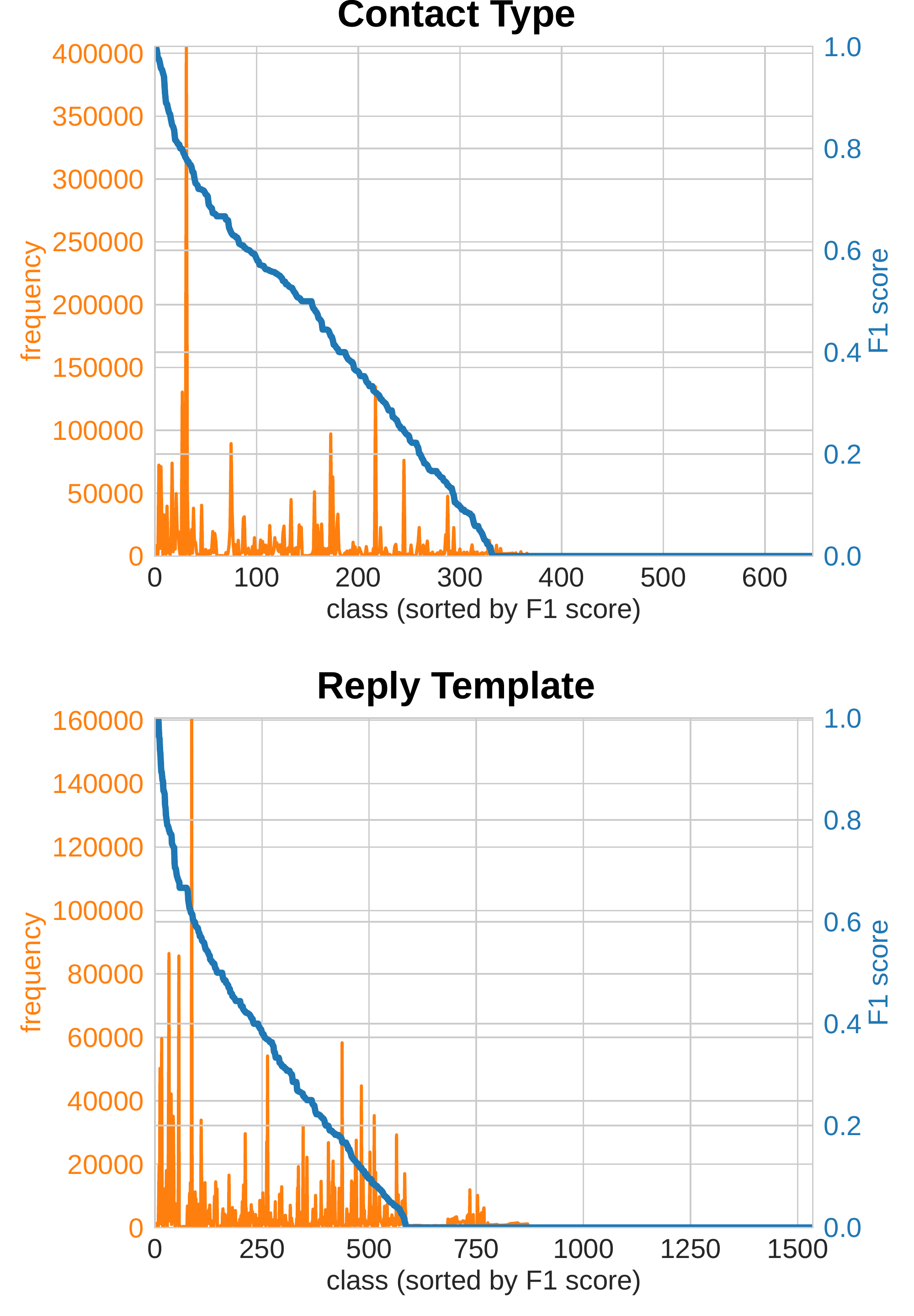}
\caption{Class frequency compared with F1 score of COTA v2 predictions on both contact types and reply templates.}
\label{fig:v2_freq_vs_f1}
\end{figure}

Class imbalance is a systematic property of the dataset: some classes (both contact types and reply templates) are rarely used by the CSRs, because those types of issues either rarely occur, or the distribution of issues the users face change continuously due to seasonality.
As a result, older classes are less relevant and rarely used.
Fig.~\ref{fig:v2_freq_vs_f1} shows how the F1 score of each class compares against the class frequency; as expected, the model performs much better on frequent classes than rare ones because there is more training data available.

The model analysis led to directions for improving the overall system, including the addition of relevant metadata as input features and the consolidation of the number of both contact types and reply templates to remove unused ones in order to have a more balanced class distribution and a higher amount of data in each class.

\subsection{Evaluation of Business Metrics}  \label{business_impact}

To measure COTA's impact, controlled A/B tests are conducted online on English language tickets.
In those experiments, there are thousands of agents, randomly assigned into either control or treatment groups.
Agents in the control group ware exposed to the original workflow without  suggestions, while agents in the treatment group are shown a modified user interface containing suggestions on contact types and reply templates produced by COTA system.
We collect tickets solved solely by either agents in the control or treatment group, and measure a few key metrics, including model accuracy, average handle time, and customer satisfaction score obtained through surveys.

The online model performance is measured and compared to offline performance.
The model performance is consistent in both the offline and online settings.
Then, customer satisfaction scores are measured and compared across control and treatment groups.
In general, customer satisfaction often increases by a few percentage points.
This finding indicates that COTA delivers the same or slightly higher quality of customer service.
Finally, to determine how much COTA affected ticket resolution speed, the average ticket handling time between the control and treatment groups is measured as well.
On average, this new feature reduced ticket handling time by ${\sim}10\%$ (p-value ${\sim}10^{-8}$).

Therefore, by injecting ML intelligence into the ticket solving process, COTA system can significantly improve agent performance and speed up ticket resolution with an improved customer satisfaction.

\section{Conclusions}

This paper describes two different model implementations of an intelligent system for improving customer support: COTA v1, a model based on feature engineering, and COTA v2, a model that exploits raw signals through deep learning architectures.
In COTA v1, a feature engineering method is employed to transform a classification task with thousands of classes to a pair-wise ranking one.
COTA v2 is based on Encoder-Combiner-Decoder, a newly proposed novel deep learning architecture that enables dealing with different inputs in a flexible way, and allows for multi-task learning.
Our experiments validate the hypotheses that 1) a ranking objective would perform better than a multi-class classification one in COTA v1, and 2) injection of prior knowledge in the form of architecture choices would improve performance making model's errors more reasonable. 
COTA v1 and COTA v2 are compared, showing how deep learning architectures perform better than a feature engineering-based architecture when learning from big datasets like the one we used.
Insights on the inner working of the model are obtained by visualizing the embeddings it learns, and its shortcomings in dealing with rare classes and imbalance in the class distribution are analyzed.
This analysis results in a set of improvements to be implemented in the future version of the system.
At the end, COTA is deployed and tested in production and the results show that it can significantly reduce ticket resolution time while improving customer satisfaction.

\begin{acks}

The authors wish to thank the Uber's Michelangelo team for the support in productionizing the models, the Opus team for the support of GPU resources, Hugh Williams, Andy Harris, Monis Ahmed Khan, Hongwei Li, Alexandru Grigoras, Viresh Gehlawat, Basab Maulik, Chinmay Maheshwari, Christina Grimsley, Chintan Shah, Franziska Bell, Taj Singh, Douglas Bemis, Felipe Petroski Such, Paul Szerlip and Eli Bingham.
\end{acks}
	
	\bibliographystyle{ACM-Reference-Format}
	\bibliography{bibliography}
	
\end{document}